\documentclass{article}
\usepackage{ijcai22}

\usepackage{times}

\usepackage{soul}
\usepackage{url}
\usepackage[hidelinks]{hyperref}
\usepackage[utf8]{inputenc}
\usepackage[small]{caption}
\usepackage{graphicx}
\usepackage{amsmath}
\usepackage{amsfonts}
\usepackage{algorithm}
\usepackage{algorithmic}
\usepackage{booktabs}
\def\ie{{\em i.e.},\ }
\newtheorem{definition}{Definition}

\title{Domain-shift adaptation via linear transformations}

\author{
Roberto Vega$^{1,2}$\and
Russell Greiner$^{1,2}$\\
\affiliations
$^1$Department of Computing Science, University of Alberta, Canada\\
$^2$Alberta Machine Intelligence Institute\\
\emails
\{rvega, rgreiner\}@ualberta.ca
}

\begin{document}

\maketitle

\begin{abstract}
A predictor, $f_A : X \to Y$, learned with data from a source domain (A) might not be accurate on a target domain (B) when their distributions are different. Domain adaptation aims to reduce the negative effects of this distribution mismatch. Here, we analyze the case where $P_A(Y\ |\ X) \neq P_B(Y\ |\ X)$, $P_A(X) \neq P_B(X)$ but $P_A(Y) = P_B(Y)$; where there are affine transformations of $X$ that makes all distributions equivalent. We propose an approach to project the source and target domains into a lower-dimensional, common space, by (1)~projecting the domains into the eigenvectors of the empirical covariance matrices of each domain, then (2)~finding an orthogonal matrix that minimizes the maximum mean discrepancy between the projections of both domains. For arbitrary affine transformations, there is an inherent unidentifiability problem when performing unsupervised domain adaptation that can be alleviated in the semi-supervised case. We show the effectiveness of our approach in simulated data and in binary digit classification tasks, obtaining improvements up to 48\% accuracy when correcting for the domain shift in the data.
\end{abstract}

\section{Introduction}
The goal of supervised machine learning is to produce a model that can accurately predict a value, $y$, given a vector input, $x$, corresponding (implicitly) to an unknown function $y=f(x)$~\cite{murphy2012machine}. In the supervised setting, we learn an approximate $\hat{y}=\hat{f}(x) \approx f(x)$, by applying a learning algorithm to a (source) training dataset $D_{S}=\{(x_1,y_1), (x_2,y_2),\dots (x_n,y_n)\}$. We can then apply this $\hat{f}$ by applying it to new instances $D_{T}=\{x_1, x_2,\dots x_m\}$.

A common assumption is that the source (S) and the target (T) domains follow the same probability distribution--i.e. $P_{S}(X,Y) = P_{T}(X,Y)$. When this is not the case, a predictor learned using $D_{S}$ might not generalize when used on $D_{T}$~\cite{storkey2009training}. The performance on the target domain depends on its performance on the source domain, and on the similarity between the distributions of the domain and target domains~\cite{ben2007analysis}.

A well known model to explain the discrepancy between distributions is covariate shift, where $P_{S}(X) \neq P_{T}(X)$, but $P_{S}(Y\ |\ X) = P_{T}(Y\ |\ X)$~\cite{shimodaira2000improving}. Other assumptions lead to different models~\cite{storkey2009training,kull2014patterns}, which motivate algorithms that decrease the negative impact of the discrepancies under different circumstances~\cite{csurka2017domain,wen2014robust}.

\begin{figure}
\centering
\includegraphics[width=\columnwidth]{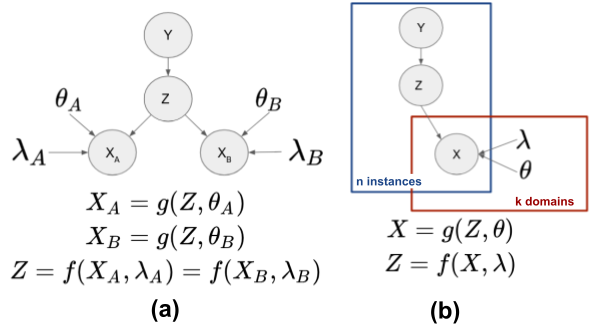} 
\caption{\textbf{(a)} Domain-shift for domains A and B. $Y$ represents the label, $Z$ represent a latent common feature space. $X_A$ and $X_B$ are the observations for the two different domains. Here, $g_{\lambda}:Z \to X$ and $f_{\theta}:X \to Z$. \textbf{(b)} Plate model for representing more than 2 domains.}
\label{fig:W_diagram}
\end{figure}

Our study focuses on the case where $P_{S} (X) \neq P_{T} (X)$,  $P_{S} (Y\ |\ X) \neq P_{T} (Y\ |\ X)$, and $P_{S} (Y) = P_{T} (Y)$. However, we assume the existence of a function $f(\cdot)$, with parameters $\lambda_A$ and $\lambda_B$, such that $P_{S} (\ f(X,\ \lambda_A)\ ) = P_{T} (\ f(X,\ \lambda_B)\ )$ and $P_{S} (\ Y\ |\ f(X,\ \lambda_A)\ ) = P_{T} (\ Y\ |\ f(X,\ \lambda_B)\ )$. This implies that there is a common feature space where the source and target domains follow the same distribution; see Figure~\ref{fig:W_diagram}. This model is called domain-shift~\cite{storkey2009training}, or covariate observation shift~\cite{kull2014patterns}.

\begin{figure}
\centering
\includegraphics[width=\columnwidth]{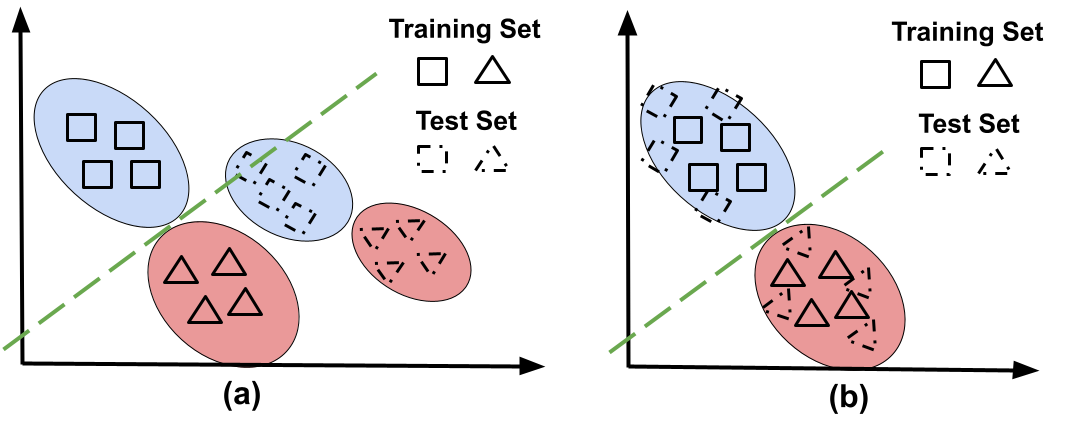} 
\caption{\textbf{(a)} The training and test set follow different probability distributions \textbf{(b)} After correcting for domain-shift, the original decision boundary now applies for both domains.}
\label{fig:simple_example}
\end{figure}

Figure~\ref{fig:simple_example}(a) exemplifies domain shift. Note that the decision boundary learned from the source domain (shown with solid squares and triangles) has poor performance on the target domain (shown with dashed squares and triangles). Domain-shift adaptation aims to find a common representation that minimizes the divergence between the domains. If successful, the decision function learned during training will have a good performance when doing inference; see Figure~\ref{fig:simple_example}(b).

Under the assumption that the mapping from source and target domains to a common representation is an affine function (\ie $Z_i = f(X_{A,i}, \lambda_A)$ and $Z_j = f(X_{B,j}, \lambda_B)$ have the form $f(X, \lambda) = \lambda X + \lambda_0$), we propose an algorithm for unsupervised and semi-supervised domain adaptation. We find the parameters $\lambda$ and $\lambda_0$, which project the data into a common space, by computing the first $p$ eigenvectors of the covariance matrices of the probability distributions of each domain, and then finding an orthogonal matrix that minimizes the maximum mean discrepancy between both distributions.

\begin{figure}
\centering
\includegraphics[width=\columnwidth]{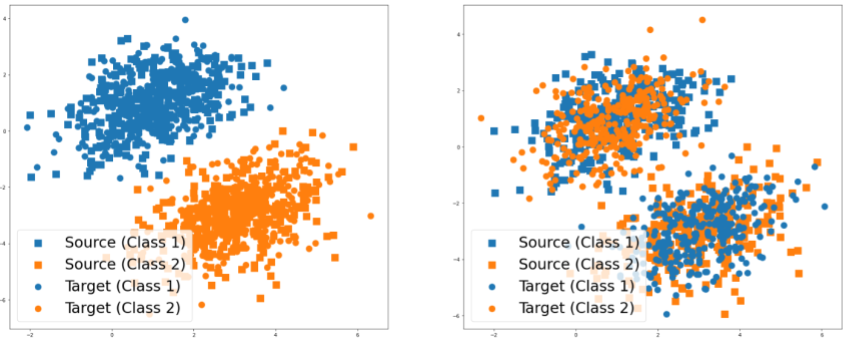} 
\caption{Anti-alignment example. In the unsupervised case it is impossible to distinguish between the scenarios of the left (correct alignment) and right (anti-alignment) graphs.}
\label{fig:anti_alignment}
\end{figure}

There is an inherent unidentifiability problem with unsupervised domain adaptation. Observe in Figure~\ref{fig:anti_alignment} that, in the absence of labeled data from the source and target domains, it is impossible to distinguish between the different ``distribution alignments'' presented there. This problem can be alleviated in the {\em semi}-supervised case, where few labeled instances allow the distinction between both scenarios.

\section{Related work}
The model presented in Figure~\ref{fig:W_diagram} is closely related to the probabilistic versions of principal components analysis~\cite{tipping1999probabilistic} and canonical correlation analysis~\cite{bach2005probabilistic}. In both cases, $Z \sim \mathcal{N}(0, I)$ and $X\ |\ Z \sim~\mathcal{N}(Wz + \mu, \Psi)$. When the transformation matrix, $W$, is diagonal --\ie a location-scale transformation-- it is possible to perform domain-shift adaptation without the Gaussianity assumption by minimizing the maximum mean discrepancy between both domains~\cite{gretton2012kernel,zhang2013domain}. In our study, we allow $W$ to be an arbitrary matrix without the Gaussianity assumption. 
Domain adaptation with arbitrary affine transformation has been explored in the context of mixing small datasets from different domains to increase the size of the training set. While this is often successful, this approach still requires a supervised dataset for each of the different sources~\cite{vega2018finding}. Here, our objective is to learn a predictor in the source domain, and then apply it to the target domain in the unsupervised and semi-supervised scenarios. 

Recent approaches attempt to minimize the divergence between source and target distributions by using data transformations. CORAL~\cite{sun2016return} matches the first two moments of the source and target distributions, while a different line of work uses variants of autoencoders to find a common mapping between source and target data~\cite{glorot2011domain,chen2012marginalized,chen2015marginalizing,louizos2016variational}. After finding this common feature space, they learned a predictor using only the source data. Their approach achieved better performance when applied to the target dataset, relative to not correcting for domain-shift, in natural language processing tasks.

Adversarial domain adaptation learns the mapping to the common space and the predictor at the same time~\cite{ganin2016domain,long2018conditional,zhao2018adversarial}. It combines a discriminator (which distinguishes instances from the different domains); a predictor (which tries to minimize the prediction error on the labeled instances); and a third function (which maps the instances into a common space). The three functions are optimized together. If successful, the discriminator should be unable to distinguish the domain on an instance (based on its common space encoding), while the predictor should have good performance under the metric of interest~\cite{tzeng2017adversarial}. 

Despite the success of neural networks for domain adaptation in natural language processing and computer vision tasks, it is hard to define what type of problems can be solved with this approach. For example, they might fail when $P_{S} (Y) \neq P_{T} (Y)$~\cite{tachet2020domain}. Even when they successfully learn an invariant representation across domains that preserves the predictive power in the source domain, this do not guarantee a successful adaptation. It is possible to have invariant representations and small error in the training set, and still have a large error in the test set~\cite{zhao2019learning}. Therefore, it is important to explicitly determine under which conditions we expect an algorithm to work. Our goal with this paper is to analyze the problem of domain-shift under affine transformations, and to propose an approach for domain adaptation for this scenario.

\section{Domain-shift adaptation via linear transformations}
Under the assumption that the domain-shift is caused by affine transformations, the equations on Figure~\ref{fig:W_diagram} become:
\begin{equation}\label{eq:assumptions}
    \begin{gathered}
    x_A = \theta_A z + \mu_A\\
    x_B = \theta_B z + \mu_B\\
    x_A \in \mathbb{R}^m,\ x_B \in \mathbb{R}^n,\ z \in \mathbb{R}^p
    \end{gathered}
\end{equation}

Note that the latent variable, $Z$, can have a different (lower) dimensionality than the observations $x_A$ and $x_B$, which in turn can have different dimensionality between themselves. Importantly, we do not assume that we have paired data between the source and target domain. In other words, for a given instance, $i$, we can either observe its representation in the source domain, $x_A^i$, or the target domain, $x_B^i$, but not both.

If we knew the parameters $\theta_A, \mu_A, \theta_B, \mu_B$,  and assuming they are non-degenerate, we could do the inverse mapping from the observations $x_A$ and $x_B$ to $z$ by solving the following optimization problem:

\begin{equation*}
    z^* = \arg \min_z ||(x - \mu) - \theta z||^2
\end{equation*}

\noindent whose solution (see the Appendix~\ref{app:optimization}) is given by

\begin{equation}\label{eq:inverse_map}
    \begin{gathered}
    z = (\theta^T \theta)^{-1}\theta^T(x - \mu)    \end{gathered}
\end{equation}

 Once we map the data from the source and target domains to a common space, we can use the labeled data from the source domain to learn a predictor that we can apply to data from any of both domains (after the appropriate projection into the common space).
 
\subsection{Estimating the transformation parameters}
Without loss of generality, we assume that $E[Z] = 0$ and $Cov[Z] = I$. Then, given a dataset with instances drawn from the source domain $X_A$ and a dataset with instances drawn from the target domain $X_B$:
\begin{equation}\label{eq:moments}
\begin{gathered}
    E[X] = \theta E[Z] + \mu = \mu\\
    Cov[X] = \theta Cov[Z] \theta^T = \theta\theta^T\\
\end{gathered}
\end{equation}

Note that we can compute the empirical estimates $\hat{\mu}_A, \hat{\mu}_B, \hat{\Sigma}_A, \hat{\Sigma}_B$, given the datasets $X_A$ and $X_B$. The empirical estimators of the mean directly give us half of the transformation parameters. For the case of the covariance matrix, we can compute the singular value decomposition:

\begin{equation}\label{eq:svd}
    \begin{aligned}
        \Sigma &= USU^T\\
        \Sigma &= US^{\frac{1}{2}}S^{\frac{1}{2}}U^T\\
        \Sigma &= US^{\frac{1}{2}}QQ^TS^{\frac{1}{2}}U^T\quad \mbox{;where} \quad QQ^T=I\\
        \Sigma &= (US^{\frac{1}{2}}Q) (US^{\frac{1}{2}}Q)^T
    \end{aligned}
\end{equation}

Since $\Sigma$ is a positive semi-definite matrix, its eigenvalues are non-negative, which allows us to decompose the diagonal matrix as $S = S^{\frac{1}{2}}S^{\frac{1}{2}}$. By comparing Equations~\ref{eq:moments} and~\ref{eq:svd}, we can estimate the parameters $\theta$ as:

\begin{equation}\label{eq:theta}
    \begin{gathered}
    \theta = U S^{\frac{1}{2}} Q
    \end{gathered}
\end{equation}

\noindent for any orthogonal matrix, $Q$. After substituting the parameter $\theta$ into Equation~\ref{eq:inverse_map}, then applying some algebraic manipulations (see the Appendix~\ref{app:rotation})
, we observe that:

\begin{equation}\label{eq:z}
    \begin{gathered}
    z = Q^T S^{-\frac{1}{2}} U^T (x - \mu)
    \end{gathered}
\end{equation}

A consequence of Equation~\ref{eq:z} is that matching the empirical mean and covariance matrices of the source and target domain is not enough to correct for domain-shift adaptation: The orthogonal matrix $Q$, which represents rotations or reflections, might cause misalignment in the data; see Figure~\ref{fig:results_Gaussian}(a).

The matrices $(S_A, U_A)$ and $(S_B, U_B)$, for the source and target domains, respectively, can be computed from the SVD of their empirical covariance matrices, $\hat{\Sigma}_A$ and $\hat{\Sigma}_B$. Since the objective of domain adaptation is to align the distributions, regardless of the ``direction'' of the alignment, we arbitrarily set $Q_A = I$. Then, we find an orthogonal matrix that minimizes the divergence between both probability distributions:

\begin{equation}\label{eq:loss}
    Q_B^* = \arg \min_{Q_B} Div(X_A\ ||\ Q_B^T X_B), \ \ s.t.\ \ Q_B Q_B^T = I
\end{equation}

\noindent where $Div(\cdot || \cdot)$ is an empirical measure of the divergence between the two domains.

\subsection{Maximum Mean Discrepancy}
A common measure of the divergence between two probability distributions is the Maximum Mean Discrepancy (MMD)~\cite{gretton2012kernel}. 

\begin{definition}[Maximum Mean Discrepancy]
Let p and q be Borel probability measures defined on a domain $\mathcal{X}$. Given observations $X := \{x_1,\dots,x_m\}$ and $Y := \{y_1,\dots,y_n\}$, drawn independently and identically distributed from p and q, respectively. Let $\mathcal{F}$ be a class of functions $f:\mathcal{X} \to \mathbb{R}$, the MMD is defined as:

\begin{equation*}
    \mbox{MMD}[\mathcal{F},p,q] := \sup_{f \in \mathcal{F}} \left( E_{x \sim p}[f(x)] - E_{y \sim q}[f(y)]\right)
\end{equation*}
\end{definition}
Informally, the purpose of the MMD is to determine if two probability distributions, $p$ and $q$, are different. The associated algorithm involves taking samples from $p$ and $q$, then finding a function that take large values on samples from $p$ and small (or negative) values on samples of $q$. The MMD is then the difference between the mean values of the function of the samples.

By defining the class of functions $\mathcal{F}$ as the unit ball in a reproducing kernel Hilbert space, \citeauthor{gretton2012kernel} (\citeyear{gretton2012kernel}) proposed (biased) empirical estimator of the $\mbox{MMD}^2$ as follows:

\begin{equation}\label{eq:MMD}
    \begin{split}
        \mbox{MMD}_b^2 [\mathcal{F}, X, Y] &= \frac{1}{m^2}\sum_{i,j=1}^m k(x_i, x_j) + \frac{1}{n^2}\sum_{i,j=1}^n k(y_i, y_j)\\
        & - \frac{2}{mn}\sum_{i,j=1}^{m,n} k(x_i, y_j)
    \end{split}
\end{equation}

\noindent where $k(\cdot, \cdot)$ is a valid kernel. In our case, we use the Gaussian kernel $k(x, y) = \exp \left(- \frac{1}{2} \sigma^{-2}||x-y||^2 \right)$.

Equation~\ref{eq:MMD} has two nice properties. (1) It computes an estimation of the MMD with a finite number of instances from each domain. (2) It is a differentiable function, so it can be optimized with iterative methods, such as gradient descent.

\subsection{Optimization with orthogonality constraints}
For solving the optimization problem with orthogonality constraints presented in Equation~\ref{eq:loss}, we used the \citeauthor{wen2013feasible} (\citeyear{wen2013feasible}) algorithm (Algorithm~\ref{alg:ortho}), which is an iterative method based on the Cayley transform. Their algorithm is similar to gradient descent, but instead of looking for solutions in the Euclidean space, they look for solutions in the Stiefel manifold, which is the set that contains all the orthogonal matrices.

Formally, their proposed algorithm solves:
\begin{equation}\label{eq:ortho}
    \min_{X \in \mathbb{R}^{n \times p}} \mathcal{F} (X), \ s.t.\ \ X^TX = I
\end{equation}

\begin{algorithm}[tb]
\caption{Optimization with orthogonality contraints}
\label{alg:ortho}
\textbf{Input}: $\mathcal{F}, X_0$\\
\textbf{Parameter}: Learning rate ($\tau$), Max iterations (M)\\
\textbf{Output}: $\arg \min_X \mathcal{F}(X)\ s.t. \ X^TX=I$\\
\begin{algorithmic}[1] 
\STATE Given an initial orthogonal matrix $X_0$.
\STATE Let $t = 0$
\WHILE{$t < M$}
\STATE Compute the Gradient $G_t = \mathcal{DF}(X_t) = \left(\frac{\partial \mathcal{F}(X_t)}{\partial X_t{i,j}} \right)$
\STATE Compute $A_t = G_t X_t^T - X_t G_t^T$
\STATE Compute $Q_t = \left(I + \frac{\tau}{2} A_t\right)^{-1}\left(I - \frac{\tau}{2} A_t\right)$
\STATE Compute $X_{t+1} = Q_t X_t$
\STATE Update $t := t+1 $
\ENDWHILE
\STATE \textbf{return} $X_M$
\end{algorithmic}
\end{algorithm}

\noindent where $\mathcal{F}: \mathbb{R}^{n \times p} \to \mathbb{R}$ is a differentiable function. For our purposes, $\mathcal{F}(Q_B) = \mbox{MMD}^2(Z_A, Q_B^T Z_B')$, where $Z_A$ and $Z_B'$ are the projections of $X_A$ and $X_B$, respectively, using Equation~\ref{eq:z} with $Q_A = Q_B' = I$. $Q_B$ is an orthogonal (rotation or reflection) matrix that multiplies $Z_B'$. 

Algorithm~\ref{alg:ortho} is guaranteed to converge when the learning rate ($\tau$) meets the Armijo-Wolfe conditions~\cite{nocedal2006numerical}. However, it is not guaranteed to find the global minimum of $\mathcal{F}(X)$. Similarly to gradient descent approaches, the algorithm might converge to a local minimum. One heuristic to alleviate this problem is to perform multiple restart with different seed points; however, this is still not guaranteed to convergence to the global minimum.

\subsection{Unsupervised domain adaptation}\label{sec:unsupervised_DA}
For finding the parameters $\theta_A$ and $\theta_B$, we can arbitrarily set $Q_A = I$ in Equation~\ref{eq:theta}, and then use Algorithm~\ref{alg:ortho}, with the Maximum Mean Discrepancy, for computing $Q_B$:

\begin{equation}\label{eq:optimal_rotation}
    Q_B^* = \arg \min_{Q_B} \mbox{MMD}_b^2(Z_A, Q_B^T Z'_B),\ s.t.\ Q_B Q_B^T = I
\end{equation}

\noindent where $Z'_B$ is a dataset that contains the transformed instances $x_B$ using $Q_B'=I$. Finally, we can project the source and target domains to a common representation using Equation~\ref{eq:z}.

Note that Algorithm~\ref{alg:ortho} requires the gradient of the $\mbox{MMD}^2$ with respect to the matrix $Q_B^T$. By applying standard matrix calculus we compute (see Appendix~\ref{app:gradient}
for details):

\begin{equation}\label{eq:gradient_optimal}
\begin{gathered}
    G(Q_B^T) = \frac{\partial \mbox{MMD}^2(Z_A, Q_B^T Z'_B)}{\partial Q_B^T} = \\
    -\frac{2}{mn} \sum_{i,j}^{n,m}\exp\left(-\frac{1}{2\sigma_2} ||z_{A}^{(i)} - Q_B^T z_{B}^{(j)}||^2\right) \left( \frac{z_{A}^{(i)} z_{B}^{(j)T}}{\sigma^2}\right)
\end{gathered}
\end{equation}

\begin{algorithm}[tb]
\caption{Unsupervised domain adaptation with linear transformations}
\label{alg:unsupervised_DA}
\textbf{Input}: $X_A \in \mathbb{R}^{i \times m}, X_B \in \mathbb{R}^{j \times n}$. Every row in these matrices represents an instance.\\
\textbf{Parameter}: Variance of Gaussian kernel ($\sigma^2$)\\
\textbf{Output}: $Z_A\in \mathbb{R}^{i \times p}, Z_B\in \mathbb{R}^{j \times p}$. Every row in each matrix represents an instance in the shared space.\\
\begin{algorithmic}[1] 
\STATE Compute $\mu_A$ and $\Sigma_A$ of the dataset $X_A$.
\STATE Compute $\mu_B$ and $\Sigma_B$ of the dataset $X_B$.
\STATE $U_A, S_A, V_A = \mbox{SVD}(\Sigma_A)$
\STATE $U_B, S_B, V_B = \mbox{SVD}(\Sigma_B)$
\STATE $\theta_A = U_A S_A^{\frac{1}{2}}$ \COMMENT{Use only the positive eigenvalues (and their corresponding eigenvectors)}
\STATE $\theta_B' = U_B S_B^{\frac{1}{2}}$ \COMMENT{Use only the positive eigenvalues (and their corresponding eigenvectors)}
\STATE $Z_A = (\theta_A^T \theta_A)^{-1}\theta_A^T (X_A - \mu_A)$
\STATE $Z_B' = (\theta_B'^T \theta_B')^{-1}\theta_B'^T (X_B - \mu_B)$
\STATE Use Algorithm~\ref{alg:ortho} to find the $Q_B^T \in \mathbb{R}^{p \times p}$ that minimizes Equation~\ref{eq:optimal_rotation}. Compute the MMD using a Gaussian kernel with variance $\sigma_2$.
\STATE $Z_B = Z_B' Q_B$
\STATE \textbf{return} $Z_A, Z_B$
\end{algorithmic}
\end{algorithm}

Algorithm~\ref{alg:unsupervised_DA} shows the procedure to map the source and target domain into a common space in an unsupervised way (The code is publicly available; see Appendix~\ref{app:code}
). For notation, the source domain (resp., target, shared space) is $m$-dimensional space, (resp. $n$-dimensional, $p$-dimensional; here we assume that $p \leq \min(m,n)$. For mapping into this lower dimensional space, we project the data into the first $p$ eigenvectors of the empirical covariance matrice $\Sigma_A$ (resp. $\Sigma_B$). The eigenvalues corresponding these eigenvectors are positive, while the other $m-p$ and $n-p$ eigenvalues will be equal to zero.

After mapping both domains to a common space, we can use the labels of the source domain to learn a predictor, and then use it to make predictions in the target domain. Section~\ref{sec:experiments} will show that the Maximum Mean Discrepancy is not convex with respect to the orthogonal matrix $Q_B^T$, meaning Algorithm~\ref{alg:ortho} might converge to a local minimum. Additionally, in the unsupervised case there is an inherent identifiability problem caused by the missing labels~\cite{david2010impossibility,koller2009probabilistic} -- \ie there are $\theta_B$ and $\theta_B'$ such that $P(\ f(X,\theta_B)\ ) = P(\ f(X, \theta_B')\ )$. This can create an ``anti-alignment'' problem; see Figure~\ref{fig:anti_alignment}. 

Note than when the ``anti-alignment'' occurs, the source and target domains have the same marginal probability $P_S(Z) = P_T(Z)$, but $P_S(Y\ |\ Z) \neq P_T(Y\ |\ Z)$. In other words, a classifier learned on the source domain will have good performance on more data from the same domain; however, it will have very poor performance on the target domain. \citeauthor{zhao2019learning} (\citeyear{zhao2019learning}) shows that aligning the marginal probability of the covariates, then learning a good predictor on the source domain, is not sufficient for successfully performing domain adaptation.

\subsection{Semi-supervised domain adaptation}
If we have access to a few labeled instances in the target domain, we might reduce the chance of converging to an ``anti-alignment''. Since the MMD is not convex with respect to the rotation matrix $Q_B^T$, a common strategy is to attempt multiple re-start (with different seed points) of an iterative optimization algorithm. We then choose the one with the lowest cost. In the unsupervised case, the MMD itself is the cost function. For the semi-supervised case, we can first run Algorithm~\ref{alg:unsupervised_DA} for each seed point, then learn a predictor, using only the labeled data from the source domain. Then, for every alignment generated by each of the seed points, evaluate those predictors on the labeled data of the target domain, and choose the one with the lowest error.

Alternatively, we could incorporate the cross entropy loss of the source and target domains into Equation~\ref{eq:optimal_rotation}. This approach requires optimizing a weighted linear combination of three terms in the loss function: the MMD, the cross-entropy in the source domain, and the cross-entropy in the target domain. Since this path requires setting these three extra weights, we limited our experiments to the first approach.

\section{Experiments and Results}\label{sec:experiments}
We first show the performance of our approach to perform domain-shift adaptation in a simulated dataset, and then, in a modified version of the MNIST digit classification task.

\subsection{Simulated data}
For the simulated data we sampled 600 instances to creaet a dataset, $D_z \in \mathbb{R}^{600 \times 2}$, from a mixture of bi-variate Gaussians with parameters $\mu_1 = \begin{bmatrix} 1 \\ 1 \end{bmatrix}$, $\mu_2 = \begin{bmatrix} 5 \\ -5 \end{bmatrix}$, $\Sigma_1 = \begin{bmatrix} 2 & 0.7 \\ 0.7 & 1 \end{bmatrix} $, $\Sigma_2 = \begin{bmatrix} 2 & 1 \\ 1 & 4 \end{bmatrix} $. These instances correspond to a common shared space $Z$. We then created two random transformation matrices $\theta_A, \theta_B \in \mathbb{R}^{5 \times 2}$, and two random translation vectors $\mu_A, \mu_B \in \mathbb{R}^5$ to create the observations. Of course, neither the real parameters, nor the instances in the shared space are visible to our algorithm.

We randomly divided $D_z$ into two disjoint datasets $D_A$ (source domain) and $D_B$ (target domain) with 300 instances each. Then, we created the datasets $X_A = D_A \theta_A^T + \mu_A$ and $X_B = D_B \theta_B^T + \mu_B$. Our algorithm only sees $X_A$ and $X_B$, which each contain 5-dimensional vectors.

\begin{figure}
\centering
\includegraphics[width=\columnwidth]{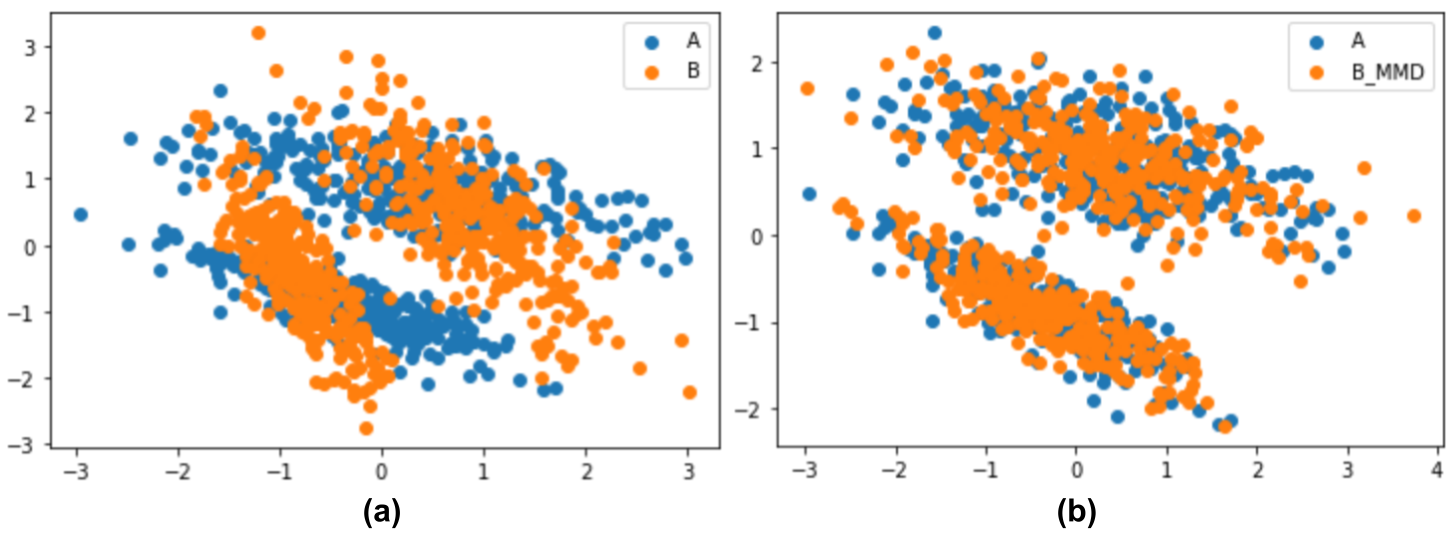} 
\caption{Results of using Algorithm~\ref{alg:unsupervised_DA} (a) without correcting for the rotation, (b) after correcting for the rotation.}
\label{fig:results_Gaussian}
\end{figure}

Figure~\ref{fig:results_Gaussian}(b) shows the result of applying Algorithm~\ref{alg:unsupervised_DA} to the simulated datasets $X_A$ and $X_B$. Figure~\ref{fig:results_Gaussian}(a), on the other hand, shows the effect of ignoring the effect of the orthogonal matrix $Q_B^T$. In this last case we successfully mapped both datasets into the same lower-dimensional space, and that both datasets have zero mean and an identity covariance matrix; however, they are not aligned. By finding the orthogonal matrix that minimizes the MMD between the source and target datasets we can obtain the correct alignment.

\begin{figure}
\centering
\includegraphics[width=\columnwidth]{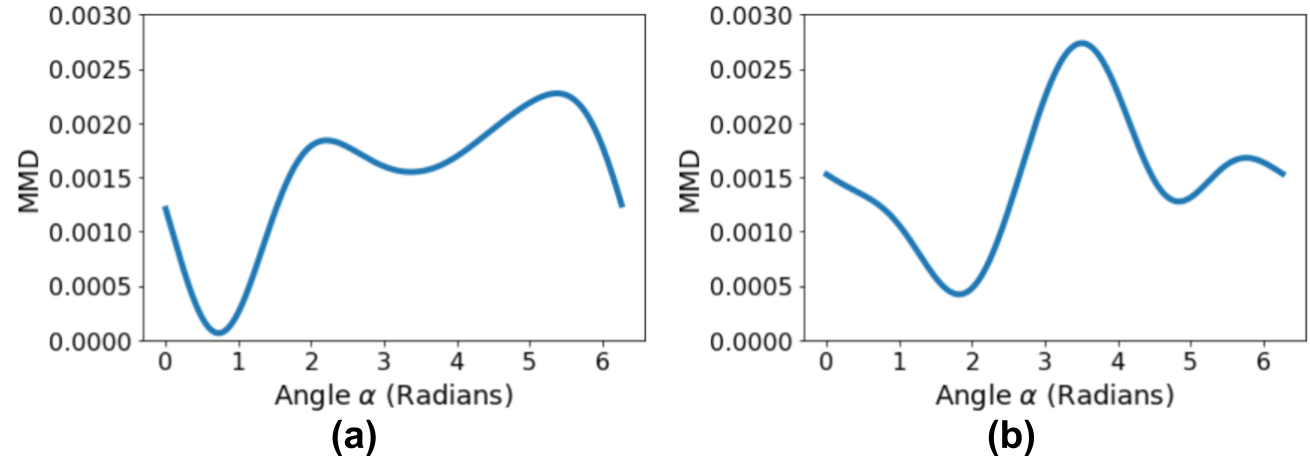} 
\caption{MMD of the simulated data for (a) rotation matrices and (b) reflection matrices.}
\label{fig:MMD_graph}
\end{figure}

As mentioned in Section~\ref{sec:unsupervised_DA}, the maximum mean discrepancy is not convex with respect to the orthogonal matrix $Q_B^T$. For 2-dimensional spaces, an orthogonal matrix is either a rotation $R = \begin{bmatrix} \cos \alpha & -\sin \alpha \\ \sin \alpha & \cos \alpha \end{bmatrix}$ or a reflection $S = \begin{bmatrix} -\cos \alpha & -\sin \alpha \\ -\sin \alpha & \cos \alpha \end{bmatrix}$~\cite{winter1992matrix}. Figure~\ref{fig:MMD_graph} shows the MMD between the projection of $X_A$ into $Z_A$ and the rotated (or reflected) projection of $X_B$ into $Z_B$ at different angles. Note that we have a total of 4 local minima for the simulated data. The global minimum corresponds to the proper alignment, shown in Figure~\ref{fig:results_Gaussian}(b). Similar to gradient descent, Algorithm~\ref{alg:unsupervised_DA} might converge to a local minimum depending on the seed point.

\begin{figure*}
\centering
\includegraphics[width=2\columnwidth]{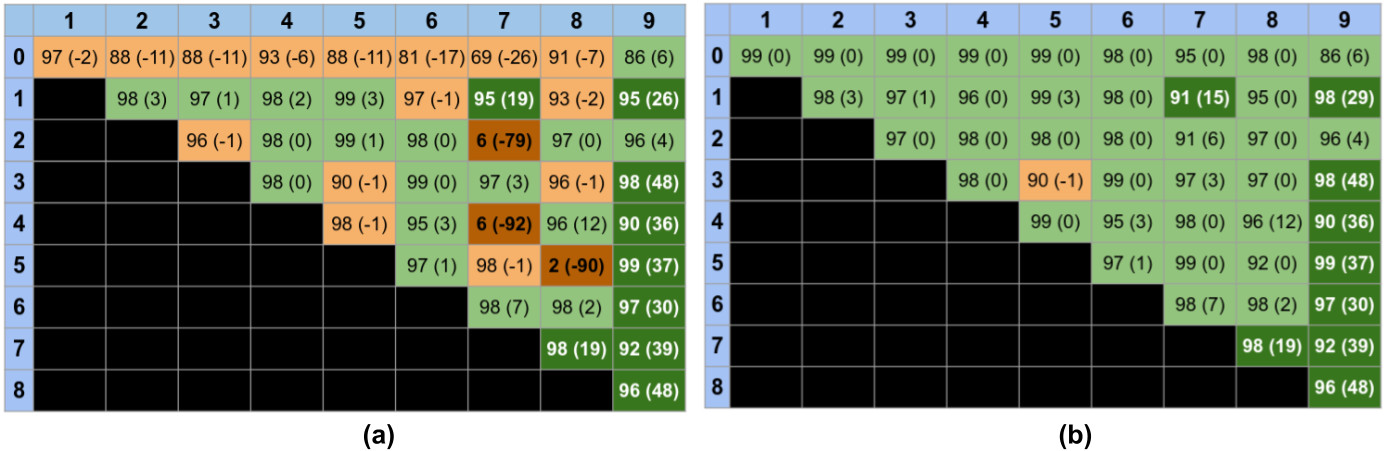} 
\caption{Results of (a) unsupervised and (b) semi-supervised domain-shift adaptation in binary digit classification.}
\label{fig:DA_Mixed}
\end{figure*}

\subsection{Binary digit classification}

The second experiment is a variation of the digit classification task with the dataset MNIST (source domain)~\cite{lecun1998gradient} and USPS (target domain)~\cite{hull1994database}. We simplified the task from 10-class digit classification to 45 binary digit classification (0 vs 1, 0 vs 2, ... , 8 vs 9).

We first trained a 10-class convolutional neural network on the training data of the source domain (60,000 images) to create image embeddings in a 20-dimensional space. The convolutional neural network contained 4 convolutional layers (32, 128, 256 and 512 filters, respectively), each followed by a MaxPooling layer ($2 \times 2$ kernel). Then we added a fully connected layer with 20 hidden neurons, and finally the output layer with 10 output neurons. While the output layer used a softmax activation function, the other layers used a rectified linear unit (ReLU) as the activation function; see Appendix~\ref{app:code}.


We used the output of the fully connected layer with 20 neurons as the image embeddings for both, the training data of the MNIST (60,000 images) and the test data USPS datasets (2,007 images). Then, for each of the 45 binary classification tasks, we compared three scenarios: (1)~Baseline: learn the parameters of a logistic regression model using the MNIST dataset, and then test it on the USPS dataset. (2)~Use Algorithm~\ref{alg:unsupervised_DA} to project the MNIST and USPS dataset into a common space of dimension 5. Then learn the parameters of a logistic regression model using only the projected data of the MNIST, and test it on the projected data of the USPS dataset. We chose a ``small'' dimensionality of the shared space because computing distances in high-dimensional spaces is harder because the instances sparsely populate the input space~\cite{friedman2001elements}. (3)~Similar to the second scenario, but now assume that 10\% of the data in the USPS dataset is labelled (semi-supervised case).

Figure~\ref{fig:DA_Mixed}(a) shows the classification accuracy of the unsupervised domain adaptation approach. The number in parenthesis indicates the improvement (or decrease) in performance relative to the Scenario 1. Similarly, Figure~\ref{fig:DA_Mixed}(b) shows the classification accuracy of the semi-supervised experiment.


\section{Discussion}
As expected, the results in Figure~\ref{fig:DA_Mixed}(a) show that even after the reducing the discrepancy between the source and target domain, and learning an accurate classifier on the source domain, this classifier is not guaranteed to generalize to the target domain. All the boxes in orange indicate that applying our algorithm for domain-shift had a lower performance than not doing any transformation at all. Specially interesting are the cases marked in dark orange, where we observe the effect of the ``anti-alignment''. On the other hand, when the alignment is done properly, there are very significant improvements in the classification accuracy. Note that for unsupervised domain adaptation there is no way to distinguish between correct and incorrect alignments.

Figure~\ref{fig:DA_Mixed}(b), on the other hand, shows that we avoid incorrect alignments in the semi-supervised case. Having access to a small set of labelled data allows the algorithm to identify when no domain-shift adaptation is needed (because the classifier already generalizes to the target domain), or detect the ``anti-alignments'' and choose the proper alignment instead; see the case of 2 vs 7, where an improper alignment occurs in the unsupervised case, but the proper alignment of the semi-supervised case increases the classification accuracy 6\%. In the case of proper alignments of the data the accuracy improved in essentially all the cases up to 48\%.

The dimensionality of the shared space plays an important role when performing domain-shift adaptation. While Figures~\ref{fig:DA_Mixed}(a) and~\ref{fig:DA_Mixed}(b) show the performance obtained in a shared space of dimension 5, Figures~\ref{fig:DA_Unsupervised_All}
and~\ref{fig:DA_Semisupervised_All}
in Appendix~\ref{app:figures}
show the performance when the dimension of the shared space is the number of positive eigenvalues in the empirical covariance matrix (13 in our experiments). The performance of the unsupervised domain adaptation degrades significantly, while the semi-supervised approach remain roughly the same. We hypothesize that this decrease in performance is due to the difficulty of reliably estimating metrics on probability distributions in high dimensional spaces with a limited number of instances~\cite{friedman2001elements}.

In summary, we present an algorithm for domain-shift adaptation caused by arbitrary affine transformations. Our approach first projects the data into a shared low-dimensional space using the first $p$ eigenvectors of the empirical covariance matrices of the data. Then, it find an orthogonal matrix that minimize the maximum mean discrepancy between the source and target data. For the unsupervised domain adaptation, there is an unavoidable identifiabiliy problem that can be alleviated by having a few labels of the target domain (semi-supervised domain adaptation). When using the correct othogonal matrix, this effectively maps both domains into a shared space where the $P(Z_A, Y) = P(Z_B, Y)$. In those cases, we can expect that a predictor learned using data from the source domain to generalize to data from a target domain.



\bibliographystyle{named}
\bibliography{ijcai22}

\begin{thebibliography}{}

\bibitem[\protect\citeauthoryear{Bach and Jordan}{2005}]{bach2005probabilistic}
Francis~R Bach and Michael~I Jordan.
\newblock A probabilistic interpretation of canonical correlation analysis.
\newblock 2005.

\bibitem[\protect\citeauthoryear{Ben-David \bgroup \em et al.\egroup
  }{2007}]{ben2007analysis}
Shai Ben-David, John Blitzer, Koby Crammer, Fernando Pereira, et~al.
\newblock Analysis of representations for domain adaptation.
\newblock {\em NeurIPS}, 19:137, 2007.

\bibitem[\protect\citeauthoryear{Chen \bgroup \em et al.\egroup
  }{2012}]{chen2012marginalized}
Minmin Chen, Zhixiang Xu, Kilian~Q Weinberger, and Fei Sha.
\newblock Marginalized denoising autoencoders for domain adaptation.
\newblock In {\em ICML}, pages 1627--1634, 2012.

\bibitem[\protect\citeauthoryear{Chen \bgroup \em et al.\egroup
  }{2015}]{chen2015marginalizing}
Minmin Chen, Kilian~Q Weinberger, Zhixiang Xu, and Fei Sha.
\newblock Marginalizing stacked linear denoising autoencoders.
\newblock {\em JMLR}, 16(1):3849--3875, 2015.

\bibitem[\protect\citeauthoryear{Csurka}{2017}]{csurka2017domain}
Gabriela Csurka.
\newblock Domain adaptation for visual applications: A comprehensive survey.
\newblock {\em arXiv preprint arXiv:1702.05374}, 2017.

\bibitem[\protect\citeauthoryear{David \bgroup \em et al.\egroup
  }{2010}]{david2010impossibility}
Shai~Ben David, Tyler Lu, Teresa Luu, and D{\'a}vid P{\'a}l.
\newblock Impossibility theorems for domain adaptation.
\newblock In {\em AISTATS'10}, pages 129--136. JMLR Workshop and Conference
  Proceedings, 2010.

\bibitem[\protect\citeauthoryear{Friedman \bgroup \em et al.\egroup
  }{2001}]{friedman2001elements}
Jerome Friedman, Trevor Hastie, Robert Tibshirani, et~al.
\newblock {\em The elements of statistical learning}, volume~1.
\newblock Springer series in statistics New York, 2001.

\bibitem[\protect\citeauthoryear{Ganin \bgroup \em et al.\egroup
  }{2016}]{ganin2016domain}
Yaroslav Ganin, Evgeniya Ustinova, Hana Ajakan, Pascal Germain, Hugo
  Larochelle, Fran{\c{c}}ois Laviolette, Mario Marchand, and Victor Lempitsky.
\newblock Domain-adversarial training of neural networks.
\newblock {\em JMLR}, 17(1):2096--2030, 2016.

\bibitem[\protect\citeauthoryear{Glorot \bgroup \em et al.\egroup
  }{2011}]{glorot2011domain}
Xavier Glorot, Antoine Bordes, and Yoshua Bengio.
\newblock Domain adaptation for large-scale sentiment classification: A deep
  learning approach.
\newblock In {\em ICML}, 2011.

\bibitem[\protect\citeauthoryear{Gretton \bgroup \em et al.\egroup
  }{2012}]{gretton2012kernel}
Arthur Gretton, Karsten~M Borgwardt, Malte~J Rasch, Bernhard Sch{\"o}lkopf, and
  Alexander Smola.
\newblock A kernel two-sample test.
\newblock {\em JMLR}, 13(1):723--773, 2012.

\bibitem[\protect\citeauthoryear{Hull}{1994}]{hull1994database}
Jonathan~J. Hull.
\newblock A database for handwritten text recognition research.
\newblock {\em IEEE Transactions on pattern analysis and machine intelligence},
  16(5):550--554, 1994.

\bibitem[\protect\citeauthoryear{Koller and
  Friedman}{2009}]{koller2009probabilistic}
Daphne Koller and Nir Friedman.
\newblock {\em Probabilistic graphical models: principles and techniques}.
\newblock MIT press, 2009.

\bibitem[\protect\citeauthoryear{Kull and Flach}{2014}]{kull2014patterns}
Meelis Kull and Peter Flach.
\newblock Patterns of dataset shift.
\newblock In {\em First International Workshop on Learning over Multiple
  Contexts (LMCE) at ECML-PKDD.}, 2014.

\bibitem[\protect\citeauthoryear{LeCun \bgroup \em et al.\egroup
  }{1998}]{lecun1998gradient}
Yann LeCun, L{\'e}on Bottou, Yoshua Bengio, and Patrick Haffner.
\newblock Gradient-based learning applied to document recognition.
\newblock {\em Proceedings of the IEEE}, 86(11):2278--2324, 1998.

\bibitem[\protect\citeauthoryear{Long \bgroup \em et al.\egroup
  }{2018}]{long2018conditional}
Mingsheng Long, Zhangjie Cao, Jianmin Wang, and Michael~I Jordan.
\newblock Conditional adversarial domain adaptation.
\newblock In {\em NeurIPS}, pages 1647--1657, 2018.

\bibitem[\protect\citeauthoryear{Louizos \bgroup \em et al.\egroup
  }{2016}]{louizos2016variational}
Christos Louizos, Kevin Swersky, Yujia Li, Max Welling, and Richard~S Zemel.
\newblock The variational fair autoencoder.
\newblock In {\em ICLR}, 2016.

\bibitem[\protect\citeauthoryear{Murphy}{2012}]{murphy2012machine}
Kevin~P Murphy.
\newblock {\em Machine learning: a probabilistic perspective}.
\newblock MIT press, 2012.

\bibitem[\protect\citeauthoryear{Nocedal and
  Wright}{2006}]{nocedal2006numerical}
Jorge Nocedal and Stephen Wright.
\newblock {\em Numerical optimization}.
\newblock Springer Science \& Business Media, 2006.

\bibitem[\protect\citeauthoryear{Shimodaira}{2000}]{shimodaira2000improving}
Hidetoshi Shimodaira.
\newblock Improving predictive inference under covariate shift by weighting the
  log-likelihood function.
\newblock {\em Journal of statistical planning and inference}, 90(2):227--244,
  2000.

\bibitem[\protect\citeauthoryear{Storkey}{2009}]{storkey2009training}
Amos Storkey.
\newblock When training and test sets are different: characterizing learning
  transfer.
\newblock {\em Dataset shift in machine learning}, 30:3--28, 2009.

\bibitem[\protect\citeauthoryear{Sun \bgroup \em et al.\egroup
  }{2016}]{sun2016return}
Baochen Sun, Jiashi Feng, and Kate Saenko.
\newblock Return of frustratingly easy domain adaptation.
\newblock In {\em AAAI}, volume~30, 2016.

\bibitem[\protect\citeauthoryear{Tachet~des Combes \bgroup \em et al.\egroup
  }{2020}]{tachet2020domain}
Remi Tachet~des Combes, Han Zhao, Yu-Xiang Wang, and Geoffrey~J Gordon.
\newblock Domain adaptation with conditional distribution matching and
  generalized label shift.
\newblock {\em NeurIPS}, 33, 2020.

\bibitem[\protect\citeauthoryear{Tipping and
  Bishop}{1999}]{tipping1999probabilistic}
Michael~E Tipping and Christopher~M Bishop.
\newblock Probabilistic principal component analysis.
\newblock {\em Journal of the Royal Statistical Society: Series B (Statistical
  Methodology)}, 61(3):611--622, 1999.

\bibitem[\protect\citeauthoryear{Tzeng \bgroup \em et al.\egroup
  }{2017}]{tzeng2017adversarial}
Eric Tzeng, Judy Hoffman, Kate Saenko, and Trevor Darrell.
\newblock Adversarial discriminative domain adaptation.
\newblock In {\em Proceedings of the IEEE conference on computer vision and
  pattern recognition}, pages 7167--7176, 2017.

\bibitem[\protect\citeauthoryear{Vega and Greiner}{2018}]{vega2018finding}
Roberto Vega and Russ Greiner.
\newblock Finding effective ways to (machine) learn fmri-based classifiers from
  multi-site data.
\newblock In {\em Understanding and Interpreting Machine Learning in Medical
  Image Computing Applications}, pages 32--39. Springer, 2018.

\bibitem[\protect\citeauthoryear{Wen and Yin}{2013}]{wen2013feasible}
Zaiwen Wen and Wotao Yin.
\newblock A feasible method for optimization with orthogonality constraints.
\newblock {\em Mathematical Programming}, 142(1):397--434, 2013.

\bibitem[\protect\citeauthoryear{Wen \bgroup \em et al.\egroup
  }{2014}]{wen2014robust}
Junfeng Wen, Chun-Nam Yu, and Russell Greiner.
\newblock Robust learning under uncertain test distributions: Relating
  covariate shift to model misspecification.
\newblock In {\em ICML}, pages 631--639. PMLR, 2014.

\bibitem[\protect\citeauthoryear{Winter}{1992}]{winter1992matrix}
David~J Winter.
\newblock {\em Matrix algebra}.
\newblock Macmillan, 1992.

\bibitem[\protect\citeauthoryear{Zhang \bgroup \em et al.\egroup
  }{2013}]{zhang2013domain}
Kun Zhang, Bernhard Sch{\"o}lkopf, Krikamol Muandet, and Zhikun Wang.
\newblock Domain adaptation under target and conditional shift.
\newblock In {\em ICML}, pages 819--827. PMLR, 2013.

\bibitem[\protect\citeauthoryear{Zhao \bgroup \em et al.\egroup
  }{2018}]{zhao2018adversarial}
Han Zhao, Shanghang Zhang, Guanhang Wu, Jos{\'e}~MF Moura, Joao~P Costeira, and
  Geoffrey~J Gordon.
\newblock Adversarial multiple source domain adaptation.
\newblock {\em NeurIPS}, 31:8559--8570, 2018.

\bibitem[\protect\citeauthoryear{Zhao \bgroup \em et al.\egroup
  }{2019}]{zhao2019learning}
Han Zhao, Remi~Tachet Des~Combes, Kun Zhang, and Geoffrey Gordon.
\newblock On learning invariant representations for domain adaptation.
\newblock In {\em ICML}, pages 7523--7532. PMLR, 2019.

\end{thebibliography}

\appendix
\section{Mathematical details}\label{app:math}
\subsection{Proof of Equation~\ref{eq:inverse_map}}\label{app:optimization}

\begin{equation*}
    \begin{aligned}
    z^* & = \arg \min_z ||(x - \mu) - \theta z||^2\\
    & = \arg \min_z \left((x - \mu) - \theta z \right)^T \left((x - \mu) - \theta z \right)\\
    & = \arg \min_z ((x - \mu)^T - z^T\theta^T) \left((x - \mu) - \theta z \right)\\
    & = \arg \min_z \left(z^T\theta^T\theta z  - 2(x - \mu)^T\theta z \right)\\
    \end{aligned}
\end{equation*}

Taking the derivative with respect to $z$ and making it equal to the zero vector:

\begin{equation*}
    \begin{aligned}
    0 &= \frac{\partial}{\partial z} \left(z^T\theta^T\theta z  - 2(x - \mu)^T\theta z \right)\\
    & = 2 \theta^T\theta z - 2\theta^T(x-\mu)\\
    \rightarrow z & = (\theta^T\theta)^{-1}\theta^T(x-\mu)
    \end{aligned}
\end{equation*}

Note that the second derivative is always non-negative, so $z$ is a minimum.

\subsection{Proof of Equation~\ref{eq:z}}\label{app:rotation}
Substituting $\theta = U S^{\frac{1}{2}}Q$ in $z = (\theta^T\theta)^{-1}\theta^T(x-\mu)$, and using that $Q^T = Q^{-1}$ for orthogonal matrices, and $(AB)^{-1} = B^{-1}A^{-1}$ for invertible matrices $A$ and $B$.:

\begin{equation*}
    \begin{aligned}
    z &= (\theta^T\theta)^{-1}\theta^T(x-\mu)\\
     &= \left( (U S^{\frac{1}{2}}Q)^T (U S^{\frac{1}{2}}Q) \right)^{-1} (U S^{\frac{1}{2}}Q)^T (x-\mu)\\
     &= \left( Q^T S^{\frac{1}{2}} U^T U S^{\frac{1}{2}}Q \right)^{-1} (U S^{\frac{1}{2}}Q)^T (x-\mu)\\
     &= \left( Q^T S^{\frac{1}{2}} S^{\frac{1}{2}}Q \right)^{-1} (U S^{\frac{1}{2}}Q)^T (x-\mu)\\
     &= \left( S^{\frac{1}{2}}Q \right)^{-1} \left( Q^T S^{\frac{1}{2}}\right)^{-1} (U S^{\frac{1}{2}}Q)^T (x-\mu)\\
     &= Q^T S^{-\frac{1}{2}} S^{-\frac{1}{2}} Q Q^T S^{\frac{1}{2}} U^T(x-\mu)\\
     &= Q^T S^{-1} S^{\frac{1}{2}} U^T(x-\mu)\\
     &= Q^T S^{-\frac{1}{2}} U^T(x-\mu)\\
    \end{aligned}
\end{equation*}

\subsection{Proof of Equation~\ref{eq:gradient_optimal}}\label{app:gradient}
For the case of the Equation~\ref{eq:MMD} using the Gaussian kernel, and since $R^TR = I$, the gradient of the MMD between $X$ and a linear transformation of $Y$, $RY$, with respect to $R$ is:

\begin{equation*}
    \begin{aligned}
        &\frac{\partial}{\partial R} \mbox{MMD}_b^2 [\mathcal{F}, X, RY] \\
        &= \frac{1}{m^2}\sum_{i,j=1}^m \frac{\partial}{\partial R} k(x_i, x_j) - \frac{2}{mn}\sum_{i,j=1}^{m,n} \frac{\partial}{\partial R} k(x_i, Ry_j) + \\
        &\ \ \quad \frac{1}{n^2}\sum_{i,j=1}^n \frac{\partial}{\partial R} k(Ry_i, Ry_j)\\
        \ \\
        &=  \frac{1}{m^2}\sum_{i,j=1}^m \frac{\partial}{\partial R} \exp \left( -\frac{1}{2\sigma^2} ||x_i - x_j||^2 \right) \\
        &\ \ \quad - \frac{2}{mn}\sum_{i,j=1}^{m,n} \frac{\partial}{\partial R} \exp \left( -\frac{1}{2\sigma^2} ||x_i - Ry_j||^2 \right) \\
        &\ \ \quad + \frac{1}{n^2}\sum_{i,j=1}^n  \frac{\partial}{\partial R} \exp \left( -\frac{1}{2\sigma^2} ||Ry_i - Ry_j||^2 \right) \\
        \ \\
        &=  - \frac{2}{mn}\sum_{i,j=1}^{m,n} \frac{\partial}{\partial R} \exp \left( -\frac{1}{2\sigma^2} (x_i - Ry_j)^T(x_i - Ry_j) \right) \\
        &\ \ \quad + \frac{1}{n^2}\sum_{i,j=1}^n  \frac{\partial}{\partial R} \exp \left( -\frac{1}{2\sigma^2} (Ry_i - Ry_j)^T (Ry_i - Ry_j) \right) \\
        \ \\
        %
        &=  - \frac{2}{mn}\sum_{i,j=1}^{m,n} \frac{\partial}{\partial R} \exp \left( -\frac{x_i^Tx_i - 2 x_i^TRy_j + y_j^Ty_j }{2\sigma^2} \right) \\
        &\ \ \quad + \frac{1}{n^2}\sum_{i,j=1}^n  \frac{\partial}{\partial R} \exp \left( -\frac{y_i^Ty_i -2y_j^Ty_i + y_j^Ty_j}{2\sigma^2}  \right) \\
        \ \\
        &=  -\frac{2}{mn}\sum_{i,j=1}^{m,n} \exp \left( -\frac{||x_i - Ry_j||^2}{2\sigma^2} \right) \left( \frac{x_i y_j^T}{\sigma^2} \right)\\
    \end{aligned}
\end{equation*}

\section{Code availability}\label{app:code}
The code for reproducing the results presented in this paper is publicly available at \url{https://github.com/rvegaml/DA_Linear}. It contains three main elements:
\begin{itemize}
    \item {\bf MLib:} An in-house developed library that contains the implementation of Algorithms~\ref{alg:ortho} and~\ref{alg:unsupervised_DA}, and auxiliary functions required to reproduce our main results.
    \item {\bf Simulations.ipynb: }A jupyter notebook with the code to reproduce the simulated experiments.
    \item {\bf BinaryDigits.ipynb: }A jupyter notebook with the code to reproduce our results with MNIST dataset.
\end{itemize}

During our experiments, we did not tune any parameter. The CNN was trained for a maximum of 500 epochs, using a learning rate of $10^{-5}$, and the default parameters of the Adam Optimizer. For the computation of the MMD we used a Gaussian kernel with $\sigma^2 = 2$.

\section{Extra figures}\label{app:figures}

\begin{figure}[h]
\centering
\includegraphics[width=\columnwidth]{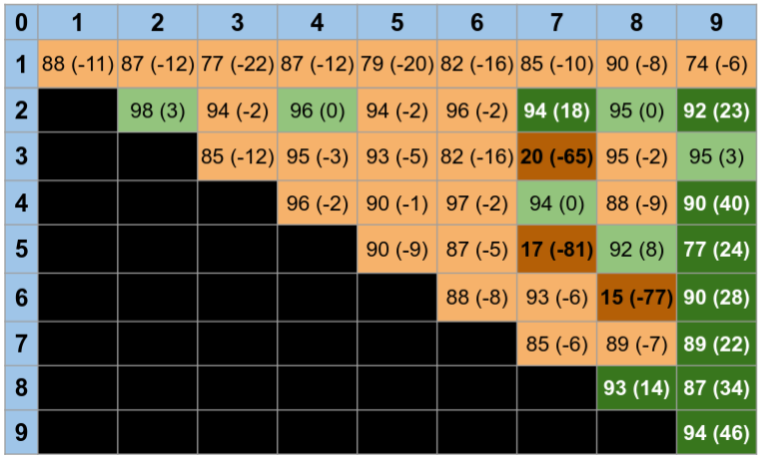} 
\caption{Results of unsupervised domain-shift adaptation in binary digit classification using a shared space of 13 dimensions.}
\label{fig:DA_Unsupervised_All}
\end{figure}

\begin{figure}[h]
\centering
\includegraphics[width=\columnwidth]{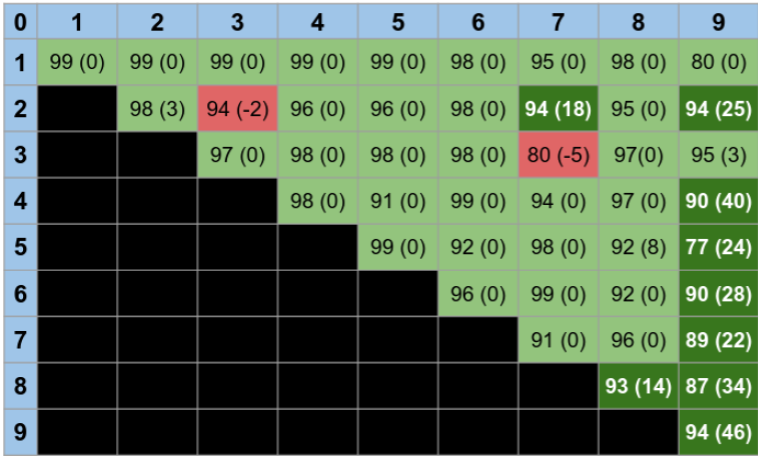} 
\caption{Results of semi-supervised domain-shift adaptation in binary digit classification using a shared space of 13 dimensions.}
\label{fig:DA_Semisupervised_All}
\end{figure}

\end{document}